# A Novel Parser Design Algorithm Based on Artificial Ants

Deepyaman Maiti[1], Ayan Acharya[2], Amit Konar[3]
Department of Electronics and Telecommunication Engineering
Jadavpur University
Kolkata: 700032, India
[1]deepyamanmaiti@gmail.com, [2]masterayan@gmail.com,
[3]konaramit@yahoo.co.in

Janarthanan Ramadoss[4]
Department of Information Technology
Jaya Engineering College
Chennai: 600017, India
[4]srmjana_73@yahoo.com

*Abstract*— This article presents a unique design for a parser using the Ant Colony Optimization algorithm. The paper implements the intuitive thought process of human mind through the activities of artificial ants. Traditional methods of designing parser involve calculation of different sets like FIRST, FOLLOW, GOTO, CLOSURE and parsing or precedence relation tables. Calculation of these tables and sets are both memory and time consuming. Moreover, the grammar concerned has to be converted into a context-free, non-redundant and unambiguous one. The scheme presented here uses a bottom-up approach and the parsing program can directly use ambiguous or redundant grammars. We allocate a node corresponding to each production rule present in the given grammar. Each node is connected to all other nodes (representing other production rules), thereby establishing a completely connected graph susceptible to the movement of artificial ants. Ants are endowed with some memory that they use to carry the sentential form derived from the given input string to the parser. Each ant tries to modify this sentential form by the production rule present in the node and upgrades its position until the sentential form reduces to the start symbol S. Successful ants deposit pheromone on the links that they have traversed through in inverse proportion of the number of hops required to complete a successful tour. Eventually, the optimum path is discovered by the links carrying maximum amount of pheromone concentration. The design is simple, versatile, robust and effective and obviates the calculation of the above mentioned sets and precedence relation tables. Further advantages of our scheme lie in i) ascertaining whether a given string belongs to the language represented by the grammar, and ii) finding out the shortest possible path from the given string to the start symbol S in case multiple routes exist.

I. INTRODUCTION

Formally, a context-free grammar is a four-tuple (*T,N,S,P*), where *T* is a set of terminal symbols, describing the allowed words, *N* is a set of non-terminals describing sequences of words and forming constructs. A unique non-terminal *S* is the start symbol. *P*, the set of production rules, describes the relationship between the non-terminal and terminal symbols, defining the syntax of the language. A series of regular expressions can be used to describe the set of allowable words, and acts as the basis for the description of a *scanner*, also called a *lexical analyzer*.

*Parsing* is the process whereby a given program is matched against the grammar rules to determine (at least) whether or not it is syntactically correct. As part of this process the various parts of the program are identified with the corresponding constructs in the grammar, so that program elements such as declarations, statements and expressions can then be identified. So, a parser for a grammar *G* is a program that takes as input a string ω and produces as output either a parse tree for ω, if ω is a sentence of G, or an error message indicating that ω is not a sentence of G.

As well as forming the front-end of a compiler, a parser is also the foundation for many software engineering tools, such as pretty-printing, automatic generation of documentation, coding tools such as class browsers, metrication tools and tools that check coding style. Automatic re-engineering and maintenance tools, as well as tools to support refactoring and reverse-engineering also typically require a parser as a front-end. The amenability of a language's syntax for parser generation is crucial in the development of such tools.

This article deals with a novel parser design algorithm based on *Ant Colony Optimization* (ACO) algorithm. The paper has been structured into 6 sections. In section II, we present a brief introduction to previous works on parsers. Section III provides a comprehensive detail of the **ACO** metaheuristic. We present our scheme in section IV. Section V highlights the advantages of our scheme. Finally, the conclusions are listed in section 6.

II. PREVIOUS WORKS ON PARSERS

Two most common forms of parsers are *operator precedence* and *recursive descent*. Two newer methods, which are more general than these and more firmly grounded in grammar theory, are: LL parsing, which really is a table-based variant of recursive descent, and LR parsing [1], [2].

The automatic generation of parsing programs from a context-free grammar is a well-established process, and various algorithms such as LL (ANTLR and JavaCC) and LALR (most notably *yacc* [3]) can be used). Application of software metrics to the measurement of context-free grammar is studied in [4]. The construction of a very wide-coverage probabilistic parsing system for natural language, based on LR parsing techniques is attempted in [5].

In [6], a design for a reconfigurable frame parser to translate radio protocol descriptions to asynchronous microprocessor cores is described. [7] presents the design and implementation of a parser/solver for semi-definite programming problems (SDPs).

[8] describes the development of a parser for the C# programming language. [9], [10] study the pattern matching capabilities of neural networks for an automated, natural language partial parser.

III. ANT COLONY OPTIMIZATION METAHEURISTICS

This section presents an overview of **ACO** algorithm which is the basis of our design. **ACO** ([11], [12], [16]) is a paradigm for designing metaheuristic algorithms for **combinatorial optimization** (**CO**) problems like Travelling Salesperson Problem ([13]), Graph Coloring Problem ([14]), Quadratic Assignment Problem ([15]) etc. The algorithms are inspired by the trail laying and following behavior of natural ants. While roaming from food sources to destination or vice versa, some of the ant species mark their paths by a chemical called **pheromone**. Other foraging ants can detect pheromone and choose, in probability, paths marked by stronger pheromone concentration. Thus pheromone trail helps the ants find the way followed by their team members towards food source or nest.

The real challenge of solving any **CO** problem by **ACO** is to map the problem to a representation that can be used by artificial ants to perform solution. In fact, any minimization problem can be represented as a triple (**S**, **f**, **Ω**) where **S** is the set of candidate solutions, **f**(s,t) is the objective function over the elements $s \in S$ and **Ω**(t) represents the problem constraints. The goal in such problems is to find a globally optimum solution $s^*$ such that $f(s^*,t) \leq f(s,t)$ for all $s \in S$. To solve such problems using the ACO metaheuristic, the problem is mapped to an environment that can be represented by connected graph $G_C=(C, L)$ ([15]), where **C**={$c_1, c_1, ...., c_N$} is the finite set of **components** and **L** is the set of links that connects fully the **components** in **C**. The states of the problem are defined in terms of sequences $x=<c_i,c_j,....,c_h,...>$ of finite length over the elements of **C. X** is the set of all possible states. The set of candidate solutions **S** is a subset of **X**. $\tilde{S}$ specifies the set of feasible candidate solution which is again a subset of **S**. the set of optimal solution $S^* \subseteq \tilde{S}$. A cost **g(s,t)** is associated with each candidate solution $s \in S$. Artificial ants build solutions by performing random walks on this construction graph to search for optimal solutions $s^* \in S^*$. Connection $l_{ij} \in L$ has associated pheromone trail $\tau_{ij}$ and a heuristic value $\eta_{ij}$. Heuristic value represents a *priori information* about the problem instance and pheromone trail conveys information to subsequent ants about the experiences gained by their predecessors.

Each ant **k** has some memory $M^k$ which is utilized for building feasible solution and retracing the path travelled backward. Ant starts from a **starting state** $x_s^k$ and build feasible solution until **termination conditions** are not met. While in state $x_r= <x_{r-1}, i>$, if termination condition is not satisfied, ant moves to a node **j** in its neighborhood $N^k(x_r)$, i.e. to state $<x_r, j>$. Choice of this node **j** is guided by a probability based selection approach given by the following equation:

$$P_i^k(j) = \begin{cases} (\tau_{ij}^\alpha).(\eta_{ij}^\beta) / \sum_{k: k \in N^k(x_r)} (\tau_{ik}^\alpha).(\eta_{ik}^\beta) \text{ if } q<q_0 \\ 1, \text{ if } (\tau_{ij}^\alpha).(\eta_{ij}^\beta) = \max\{(\tau_{ik}^\alpha).(\eta_{ik}^\beta): k \in N^k(x_r)\} \text{ with } q>q_0 \\ 0, \text{ if } (\tau_{ij}^\alpha).(\eta_{ij}^\beta) \neq \max\{(\tau_{ik}^\alpha).(\eta_{ik}^\beta): k \in N^k(x_r)\} \text{ with } q>q_0 \end{cases} \quad (1)$$

with $P_i^k(j)$ is the probability of selecting node *j* after node *i* for ant *k*. $N^k(x_r)$ being the neighborhood of ant *k* when it is at node *i* or in other words in state $x_r$). $0<q_0<1$ is a pseudo random factor deliberately introduced for path exploration. *q*, a random number generated every time ant updates its position, also lies between 0 and 1. *α*, *β* are the weights for pheromone concentration and visibility. After building a solution successfully, ant can retrace its path and deposit pheromone on the links that it has traversed through.

Therefore, ACO algorithm can be thought of as interplay of three procedures as depicted in the following pseudo code.

**Procedure** ACO metaheuristic

    **Schedule Activities**

        **Construct Solution**
        **Update pheromone**
        **Daemon Actions** (optional)
    **end Schedule Activities**

**end Procedure**

IV. OUR ALGORITHM

This section describes our design in detail. Let ω = abbcde be a string which may or may not belong to L(G), the set of strings identified by the grammar G. The given production rules are:

1. S → aAcBe ,   2. A → Ab,   3. A → e

4. $A \rightarrow b$, 5. $B \rightarrow Bdc$, 6. $B \rightarrow d$

The parser will check if ω can be reduced to S (the start symbol) by using the production rules. The quickest process to verify that is to use the production rules in the sequence shown below:

$$\omega = abbcde$$
$$\rightarrow abbcBe \quad (\text{Rule 6}), \rightarrow aAbcBe \quad (\text{Rule 4}),$$
$$\rightarrow aAcBe \quad (\text{Rule 2}), \rightarrow S \quad (\text{Rule 1})$$

So, the parser program changes sub-strings of ω which matches the RHS of a production rule by the LHS of that production rule to get a new string. In this way it continues until a new string is obtained that matches/ is the start symbol S. If ultimately S is obtained as a new string, $\omega \in L(G)$, else it is not.

Now, suppose, we are given to check whether the string ω = abbcde belongs to *L*(G) or not under the given production rules (1) - (6), i.e. we have to check whether ω can be reduced to *S* or not using the given production rules. While implementing ACO algorithm to solve this problem, we first map the entire problem into an environment represented by a connected graph. This is, as already mentioned in section 3, the first step of solving any optimization problem by a discrete optimization algorithm like ACO.

Corresponding to each production rule, which we can access to reduce ω to S, we create a node. Every node stores a production rule after splitting it into two halves. One is the LHS of the rule and the other is its RHS. Suppose, rule (1): $S \rightarrow aAcBe$ is needed to be stored in a node. Then, first we split the rule into two halves and store the LHS part S and the RHS part aAcBe separately. Therefore, each node contains exactly two stacks for storing the two halves of the production rule that it represents. A single node is connected to every other node, which implies that there is a provision of traversing from one node to any other node.

Artificial ants are endowed with some attributes to move through this connected graph. Each ant is provided with the starting string ω. This string is stored in a stack in ant's memory. In the inception, ants are placed randomly on nodes and each ant tries to use the stacks already stored in the node in which it is placed. The procedure can be illustrated using the problem string "abbcde" and the rule $B \rightarrow d$ (rule 6). An ant placed in the node representing the rule $B \rightarrow d$ starts with the expression "abbcde" and tries to find whether any sub-string of "abbcde" matches with the string corresponding to the RHS expression "d". If there is a match, the ant replaces the sub-string of ω = abbcde with the other string present in the node (LHS of the production rule). Here an ant finds that the string "d" matches with a sub-string stored in its memory. Therefore it modifies the string stored in its memory to "abbcBe".

Ant's transition from one node to another is guided by a probability based selection approach given in the following equation:

$$P_i^k(j) = \begin{cases} (\tau_{ij}) / \sum_{k: k \in N_i^k} (\tau_{ik}) \text{ if } q < q_0 \\ 1, \text{ if } \tau_{ij} = \max\{(\tau_{ik}): k \in N_i^k\} \text{ with } q > q_0 \\ 0, \text{ if } \tau_{ij} \neq \max\{(\tau_{ik}): k \in N_i^k\} \text{ with } q > q_0 \end{cases} \quad (2)$$

The difference between (1) and (2) is that here we do not consider any heuristic information. In the beginning, the search space is covered with uniform pheromone concentration. So, ants do not have any idea as how to move through the graph. It, therefore, selects the next node randomly and tries to match the modified string that it contains now in its memory with the strings stored in that node. If ant finds the strings useful so that either of them matches with any sub-string of the expression stored in ant's memory, ant modifies the expression with the other string stored in that node as before and moves to the next node. If an ant does not find the information in the current node useful, it stops moving in the graph (i.e. it becomes inactive). Each time an ant updates its position (i.e. moves from one node to the other), it checks the expression that it is modifying and the expression that it has to arrive at (which is S). If these strings match, it indicates that the ant has discovered steps through which *S* can be arrived at starting from ω and the ant stops moving. There is, however, no restriction in visiting a particular node more than once because in process of deriving one expression from the other, we might require to use a rule more than once.

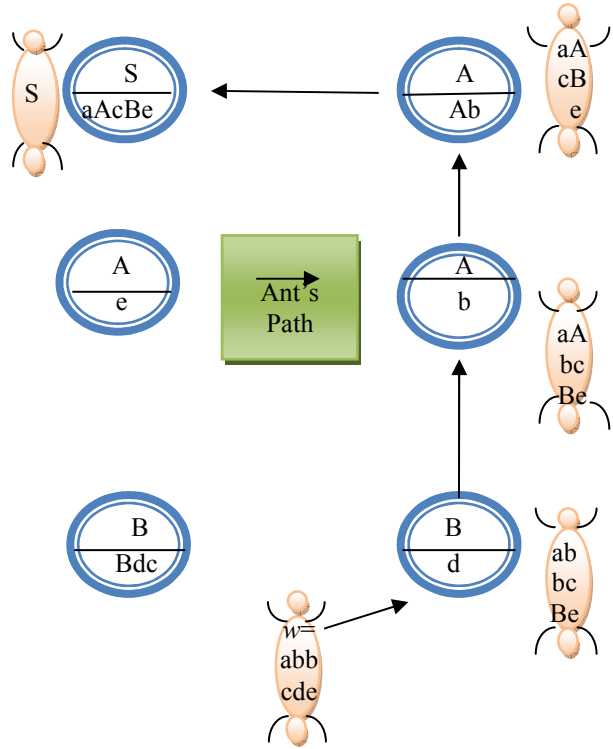

**Figure 1. Pictorial representation of our scheme**

But an ant counts the number of hops it requires to move from the starting node to the ending node and deposits pheromone on the links in inverse proportion of the number of hops. It implies that ant, which arrives at S from the starting expression using minimum number of rules, deposits maximum amount of pheromone on the links that it has traversed through. As the algorithm progresses, only a few links, which are conducive in guiding the ants towards an optimal solution, receive increasing amount of pheromone which eventually leads to the exploration of the shortest possible steps to check the validity of the given string. Figure (1) shows a visual representation of the entire scheme.

## V. Advantages of Proposed Scheme

This section highlights the advantages of our scheme over all the existing algorithms of designing a parser. The benefits are summarized below:

1. We do not need to use up resources for calculating FIRST, FOLLOW, GOTO or CLOSURE sets or parsing or precedence relation tables which are required by more advanced types of parsers. Nor do we require to get rid of back-tracking. Since there are numerous ants, the appropriate production rule to be used is found easily. So we do not need to left-factor to eliminate back-tracking, since the elimination of back-tracking is not needed at all. We can directly work with ambiguous grammars.
2. There is no need to eliminate left-recursion either, since one ant will invariably find the correct path; and when one does, the others will soon follow suit. So we do not require working with a context-free or a non-redundant grammar. This is a huge simplification to the parsing problem.
3. Our scheme is based on the intuitive method of human thinking, and thus conceptually simpler, and easier to visualize.

While it is true that the scheme 'could' have been implemented by exploring paths from ω to S randomly, the use of a stochastic optimization strategy, namely ACO is justified due to the following two conditions:

i) if ω is NOT a legal string of L(G), it is difficult to ascertain the fact without using a systematic search policy.
ii) using an optimization algorithm gives the shortest route from ω to S; this is essential to increase the speed of operation of the parser, especially when there are multiple routes, which is the general case, and the usual case for complex applications.

## VI. Conclusions and Scope of Future Work

The many advantages of the proposed parsing scheme point towards the fact that this approach will be suitable for parsing complex expressions, such as those encountered in natural language analysis applications. We use the very basic bottom-up approach, so the scheme is conceptually simple. The use of the ACO metaheuristic ensures that we can use ambiguous and redundant grammars. In the future, we plan to use the ACO algorithm to design more advanced parser types.

## References

[1] A. Aho, R. Sethi, J. Ullman, *Compilers: Principles, Techniques and Tools,* Addison-Wesley, Reading, Massachusetts, USA, 1986.
[2] E. Black, R Garside, G. Leech, *Statistically driven computer grammars of English: the IBM/Lancaster approach*, Rodopi, 1993.
[3] S. C. Johnson, "YACC – yet another compiler compiler", *Computer Science Technical Report 32*, AT&T Bell Laboratories, Murray Hill, NJ, USA, 1975.
[4] J. Power, B. Malloy, "Metric-based analysis of context-free grammars", *8th Int Workshop on Program Comprehension*, IEEE Computer Society, pp 171-178, Limerick, Ireland, 2000.
[5] T. Briscoe, J. Carroll, "Generalized probabilistic LR parsing of natural language (Corpora) with Unification-Based Grammars", *Computational Linguistics, vol. 19, issue* 1, pp. 25- 59, 1993.
[6] D. Guha, T. Srikanthan, "Reconfigurable frame parser design for multi-radio support on asynchronous microprocessor cores", *Int Conf on Computing: Theory and Applications*, pp 122-127, 2007.
[7] W. Shao-Po, S. Boyd, "Design and implementation of a parser/solver for SDPs with matrix structure", *1996 IEEE Int Symposium on Computer-Aided Control System Design*, pp 240-245, 1996
[8] B. A. Malloy, J. T. Waldron, "Applying software engineering techniques to parser design: The Development of a C# Parser", *Annual Conf of the South African Institute of Computer Scientists and Information Technologists*, pp 75-82, 2002.
[9] H. T. Siegelmann, C. L. Giles, "The complexity of language recognition by neural networks", *Neurocomputing*, vol. 15, issues 3-4, pp 327-345, 1997.
[10] C. Lyon, R. Frank, "Neural network design for a natural language parser", *International Conf on Artificial Neural Networks*, pp 105-110, 1995
[11] M. Dorigo, C. Blum. "*Ant colony optimization theory: A survey*", Theoretical Comp. Sc. 344, pp. 243–278, 2005
[12] M.Dorigo, T.Stiizle, *Ant colony optimization*, MIT Press, Cambridge, MA, 2004.
[13] M.Dorigo, L.M.Gambardella, "Ant colonies for the traveling salesman problem," *Bio Systems*, vol. 43, no.2, pp. 73-81, 1997.
[14] D.Costa, A.Hertz, "Ants can color graphs", *Journal of the Operational Research Society*, vol. 48, no. 3, pp. 295-305, 1997.
[15] V. Maniezzo, A. Colorni, "The ant system applied to the quadratic assignment problem", *IEEE Trans. on Knowledge and Data Engineering* vol. 11, issue 5, pp. 769 – 778, 1999.
[16] M. Dorigo, M. Birattari, T. Stiitzle, "Ant colony optimization: artificial ants as a computational intelligence technique," *IEEE Computational Intelligence Magazine*, vol. 1, no. 4, 2006.